\documentclass{article} 
\usepackage{iclr2027_conference,times}
\usepackage[utf8]{inputenc}

\usepackage{amsmath,amsfonts,bm}

\def\eqref#1{equation~\ref{#1}}

\def\1{\bm{1}}

\DeclareMathAlphabet{\mathsfit}{\encodingdefault}{\sfdefault}{m}{sl}
\SetMathAlphabet{\mathsfit}{bold}{\encodingdefault}{\sfdefault}{bx}{n}

\DeclareMathOperator*{\argmin}{arg\,min}

\usepackage{hyperref}
\usepackage{url}
\usepackage{graphicx}
\usepackage{float}
\usepackage{booktabs}
\usepackage{multirow}
\usepackage{amsmath,amssymb}
\usepackage{titlesec}
\titlespacing*{\subsection}{0pt}{6pt plus 1pt minus 1pt}{3pt}
\titlespacing*{\section}{0pt}{8pt plus 1pt minus 1pt}{4pt}
\usepackage{subcaption}
\usepackage{tikz}
\usetikzlibrary{arrows.meta,positioning,shapes.geometric,calc}
\tikzset{
  box/.style={draw, rectangle, minimum height=0.6cm, minimum width=1.7cm, align=center, font=\small, inner sep=3pt},
  dropbox/.style={draw, rectangle, dashed, minimum height=0.6cm, minimum width=1.7cm, align=center, font=\small, inner sep=3pt, fill=black!8},
  decision/.style={draw, diamond, aspect=2.4, align=center, font=\scriptsize, inner sep=1pt},
  note/.style={draw, rectangle, align=left, font=\scriptsize, inner sep=5pt, text width=0.85\linewidth},
  arr/.style={-{Stealth[length=2mm]}, thick},
}

\title{A Rigor-Matched Audit of Periodic-Step Layer Skipping for Efficient LLM Inference: ConfLayers versus SWIFT, with a Supplemental Analysis of Trained Routing Alternatives}

\author{Prateek Kumar Sikdar \\
\texttt{prateek.k.sikdar@accenture.com}
}

\newcommand{\layerroute}{LayerRoute}
\newcommand{\layerdrop}{LayerDrop}
\newcommand{\conflayers}{ConfLayers}
\newcommand{\swift}{SWIFT}

\iclrfinalcopy
\begin{document}

\maketitle
\thispagestyle{fancy}
\fancyhead{}
\lhead{}

\begin{abstract}
Layer-skipping methods for efficient LLM inference decide, at some granularity, which transformer layers to execute for a given input. We present a rigor-matched, three-seed audit of two \emph{periodic-step, search-based} methods that make this decision online, at inference time, and re-evaluate it every few generation steps: a confidence-gated early-exit baseline (\conflayers{}) and genuine self-speculative decoding (\swift{}, \citealt{xia2024swift}), together with vanilla autoregressive decoding, across two model scales (Qwen2.5-0.5B and Qwen2.5-1.5B, \citealt{yang2024qwen2}) and two tasks (GSM8K reasoning, \citealt{cobbe2021gsm8k}; CNN/DailyMail summarization, \citealt{nallapati2016cnndm,see2017cnndm}). \swift{} is the strongest method on accuracy in three of four cells; \conflayers{} is dominated everywhere, with particularly large deficits on GSM8K at 1.5B. Once online-search overhead is correctly separated from pure inference cost --- a decomposition we introduce and validate --- \swift{}'s true inference speed is faster than \conflayers{}'s in all four cells (5--21\%), reversing the naive wall-clock ranking in three of them; \conflayers{}'s search overhead is small and stable (1--2\% of cost) while \swift{}'s is larger and considerably more variable seed-to-seed (up to 28.7\%). We additionally examine two \emph{trained-routing} methods, \layerroute{} \citep{sikdar2026layerroute} (a per-sequence, input-conditioned hard gate) and \layerdrop{} \citep{fan2020layerdrop} (a fixed, input-independent pruning pattern), as a supplemental analysis rather than a head-to-head comparison, since both operate at a fundamentally coarser decision granularity than the periodic-step methods above. Measuring both under a verified protocol --- genuine per-input gating, a genuine full-model baseline, and genuine inference-time compute skipping --- both trained-routing methods show modest, real speedups (1.08--1.33$\times$) but accuracy well below the periodic-step methods, including a near-total collapse for \layerroute{} on GSM8K at 1.5B (0.003 mean exact-match across three seeds). We release the full audit protocol as a template for rigor-matched efficiency comparisons.
\end{abstract}

\section{Introduction}
\label{sec:intro}

Layer-skipping methods for efficient LLM inference share a common shape --- decide, for some unit of input, whether to execute a given transformer layer or bypass it --- but differ sharply in \emph{when} and \emph{at what granularity} that decision is made. Confidence-gated early-exit methods threshold a per-token confidence signal that is itself tuned online \citep{schuster2022calm,schuster2021consistent,teerapittayanon2016branchynet}; self-speculative decoding methods draft candidate continuations with a cheapened version of the same model, periodically re-evaluating the draft configuration, and verify drafts against the full model in a single batched pass \citep{leviathan2023speculative,chen2023speculative,xia2024swift}. Both families make their skip decision \emph{online}, re-checking it every few generation steps. A separate family, \emph{trained routing}, instead learns a fixed policy offline --- a gate, a router, or a pruning mask --- so no online search is needed \citep{fan2020layerdrop,raposo2024mixture,elbayad2020depth}; the policy can be as coarse as a single fixed pattern applied to every input, or as fine as a decision recomputed per sequence.

This granularity difference determines what a fair comparison even means: a method re-evaluating its configuration every 25--30 tokens is answering a different question than one committing to a single decision for an entire sequence, or one that never adapts to input at all. Our primary contribution is therefore a comparison restricted to methods sharing the same decision regime --- \conflayers{} and \swift{}, both periodic-step and search-based --- audited under matched seeds, tasks, and scales, with a cost-accounting methodology that correctly separates the online search from the inference it accelerates. This confound is methodologically significant: once search overhead is excluded, the pure-inference speed ranking of the two methods reverses relative to their naive wall-clock ranking in three of four task/scale combinations we test.

We additionally examine two representative trained-routing methods, \layerroute{} \citep{sikdar2026layerroute} and \layerdrop{} \citep{fan2020layerdrop}, as a \emph{supplemental analysis} rather than folding them into the main comparison, since their decision granularity does not match \conflayers{}'s and \swift{}'s: \layerroute{} commits to one gate decision per sequence; \layerdrop{} commits to one fixed pattern that never varies with input. Treating either as a direct peer of a method that re-searches every 25 tokens would overstate what the comparison can claim. We measure both under a protocol verified to exercise their routing decisions genuinely --- real per-input gating, a real full-model baseline, and real inference-time compute skipping --- and report the resulting numbers plainly, including a result we did not expect going in.

Our contributions are:

\begin{itemize}
\setlength{\itemsep}{1pt}
\setlength{\parskip}{0pt}
\item A rigor-matched, 3-seed, 2-task $\times$ 2-scale comparison of two periodic-step, search-based layer-skipping methods (\conflayers{}, \swift{}) against vanilla decoding, with a taxonomy of routing decision granularity motivating why this is the correct main comparison set (Section~\ref{sec:taxonomy}).
\item A search-overhead decomposition that separates each method's pure inference cost from its online-search cost, revealing that naive wall-clock comparisons had inverted the true inference-speed ranking of \swift{} and \conflayers{} in three of four cells (Section~\ref{sec:overhead}).
\item A supplemental analysis of two trained-routing methods (\layerroute{}, \layerdrop{}) at a different decision granularity, using a measurement protocol verified to exercise genuine per-input gating, a genuine full-model baseline, and genuine inference-time compute skipping throughout (Section~\ref{sec:supplemental}).
\item A reusable, rigor-matched audit protocol, including the frozen-replay search-overhead decomposition, applicable to future efficiency comparisons.
\end{itemize}

\section{A Taxonomy of Layer-Skipping Decision Granularity}
\label{sec:taxonomy}

Table~\ref{tab:taxonomy} summarizes the four methods discussed in this paper along the dimension that we argue determines comparability: how often, and on what unit of input, the skip decision is made.

\begin{table}[!h]
\caption{Layer-skipping methods by decision granularity. \conflayers{} and \swift{} share the same regime (periodic-step, search-based) and form our main comparison; \layerroute{} and \layerdrop{} occupy two different, coarser regimes and are reported as a supplemental analysis rather than head-to-head peers.}
\label{tab:taxonomy}
\begin{center}
\small
\begin{tabular}{lccc}
\toprule
Method & Training & Decision unit & Re-evaluation cadence \\
\midrule
\conflayers{} & None (online search) & Per token & Every 30 steps \\
\swift{} & None (online search) & Per token & Every 25 steps \\
\midrule
\layerroute{} & Trained (LoRA + gate) & Per sequence & Once, input-conditioned \\
\layerdrop{} & Trained (structured dropout) & N/A & Never (fixed, input-independent) \\
\bottomrule
\end{tabular}
\end{center}
\end{table}

\conflayers{} and \swift{} both decide online, re-searching their configuration every 25--30 generation steps and often converging early, freezing the found configuration for the remainder of a session (Section~\ref{sec:overhead}); this shared regime is what makes a matched comparison between them meaningful. \layerroute{} is designed to make one gate decision per input sequence, using that sequence's own hidden states --- a coarser, input-conditioned regime that, absent an explicit freezing mechanism, can drift toward continuously-updating behavior in genuine generation (Section~\ref{sec:supplemental-layerroute}). \layerdrop{} is coarser still: its pruning pattern is fixed at training time and applied identically to every input, with no adaptivity at all. Comparing \conflayers{}/\swift{} against \layerroute{}/\layerdrop{} on identical footing would compare methods answering different questions; we instead present the latter two as a supplemental analysis with this distinction stated explicitly.

\section{Related Work}
\label{sec:related}

\textbf{Trained routing and structured skipping.} \layerdrop{} \citep{fan2020layerdrop} trains transformers to be robust to layer removal via structured dropout, enabling depth reduction at inference without retraining; we evaluate it directly in Section~\ref{sec:supplemental}. The Depth-Adaptive Transformer \citep{elbayad2020depth} and PonderNet \citep{banino2021pondernet} extend Adaptive Computation Time \citep{graves2016adaptive} to learn per-example halting policies. Mixture-of-Depths \citep{raposo2024mixture} routes a fixed fraction of tokens per layer through a top-$k$ router, a finer (token-level) granularity than \layerroute{}'s per-sequence gate; we do not evaluate it here. Post-hoc depth-pruning work such as ShortGPT \citep{men2024shortgpt} and \citet{gromov2024unreasonable} establish that later transformer layers often contribute disproportionately little to output quality, independent of any routing mechanism.

\textbf{Confidence-gated and search-based early exit.} CALM \citep{schuster2022calm} and its earlier variant \citep{schuster2021consistent} exit a token's forward pass once a per-layer confidence measure crosses a threshold; BranchyNet \citep{teerapittayanon2016branchynet} applies the same idea to convolutional networks. Our \conflayers{} baseline follows this family, augmented with an online Bayesian-optimization search over per-layer thresholds (Section~\ref{sec:setup}).

\textbf{Speculative and self-speculative decoding.} Speculative decoding \citep{leviathan2023speculative,chen2023speculative,stern2018blockwise} drafts multiple tokens with a cheap model and verifies them in one batched pass, with later work exploring tree-structured verification \citep{miao2024specinfer,cai2024medusa} and lightweight adapters as drafters \citep{du2024glide}. \swift{} \citep{xia2024swift} removes the need for a separate draft model by self-drafting with a layer-skipped version of the target model, discovered online rather than trained offline.

\textbf{Compute-efficient scaling and pruning.} Our discussion of scale-dependent behavior connects to compute-optimal scaling \citep{hoffmann2022training} and to structural pruning such as LLM-Pruner \citep{ma2023llmpruner} and Wanda \citep{sun2024simple}, which similarly find removable redundancy is scale-dependent.

\section{Background: Method Formalizations}
\label{sec:background}

We formalize \conflayers{} and \swift{} here; \layerroute{} and \layerdrop{}'s formalizations appear in the supplemental analysis (Section~\ref{sec:supplemental}).

\subsection{The ConfLayers Baseline}
\label{sec:background-conflayers}

\conflayers{} \citep{amer2026conflayers} requires no training: it wraps a fixed pretrained backbone with a per-layer confidence rule and an online search over its thresholds.

\textbf{Confidence signal and exit rule.} At layer $l$, a confidence estimate projects the current hidden state to the vocabulary and takes the top probability,
\begin{equation}
c_l(h_l) = \max_{v} \; \mathrm{softmax}\!\left(W_{lm}\, h_l\right)_v,
\label{eq:conf}
\end{equation}
where $W_{lm}$ is the shared, frozen LM head. The model exits at the first layer clearing threshold $\tau_l$,
\begin{equation}
z_l = \mathbb{1}\!\left[c_l(h_l) \geq \tau_l\right], \qquad \text{exit at } l^* = \min\{\, l : z_l = 1 \,\},
\label{eq:conf-exit}
\end{equation}
emitting $W_{lm} h_{l^*}$ directly as output logits, skipping all $l > l^*$ (Figure~\ref{fig:arch-conflayers}).

\textbf{Online threshold search.} $\tau = (\tau_1, \dots, \tau_L)$ is not learned by gradient descent; since the mapping from $\tau$ to quality and latency is a black-box, non-differentiable function, $\tau$ is tuned online via Bayesian optimization,
\begin{equation}
\tau^\star = \argmin_{\tau \in [0,1]^L} \; \mathbb{E}_x\!\left[\mathrm{Cost}(x; \tau)\right] \quad \text{s.t.} \quad \mathbb{E}_x\!\left[\mathrm{QualityDrop}(x; \tau)\right] \leq \epsilon,
\label{eq:conf-search}
\end{equation}
proposing candidate $\tau$ vectors, evaluating on a held-out probe set, and updating a surrogate cost--quality model. This search is what we isolate as ``search overhead'' in Section~\ref{sec:overhead}: wall-clock time spent choosing $\tau$, separate from the forward-pass cost of Equations~\ref{eq:conf}--\ref{eq:conf-exit} once fixed.

\subsection{The SWIFT Baseline}
\label{sec:background-swift}

\swift{} \citep{xia2024swift} also requires no training, but rather than exiting early it drafts multiple tokens with a cheapened version of the target model and verifies them losslessly against the full model.

\textbf{Self-draft model.} Given a skip set $S \subseteq \{1, \dots, L\}$, the self-draft model $\mathcal{M}_S$ is $\mathcal{M}$ with every layer $l \in S$ replaced by an identity pass-through,
\begin{equation}
h_l^{(S)} = \begin{cases} h_{l-1}^{(S)} & l \in S \quad \text{(skipped)} \\ f_l\!\left(h_{l-1}^{(S)}\right) & l \notin S \quad \text{(executed)} \end{cases},
\label{eq:swift-draft-model}
\end{equation}
requiring no additional parameters, since $\mathcal{M}_S$ reuses $\mathcal{M}$'s own weights (Figure~\ref{fig:arch-swift}).

\textbf{Draft-then-verify.} $\mathcal{M}_S$ autoregressively drafts $k$ candidate tokens, and $\mathcal{M}$ verifies all $k$ positions in one batched pass. Each draft token is accepted with probability
\begin{equation}
\min\!\left(1, \; \frac{P_{\mathcal{M}}(\tilde{x}_{t+i} \mid x_{\leq t+i-1})}{P_{\mathcal{M}_S}(\tilde{x}_{t+i} \mid x_{\leq t+i-1})}\right),
\label{eq:swift-accept}
\end{equation}
following the standard rejection rule \citep{leviathan2023speculative}; at the first rejection, a corrective token is resampled from $p_{\mathrm{res}}(x) \propto \max\!\left(0,\, P_{\mathcal{M}}(x) - P_{\mathcal{M}_S}(x)\right)$, guaranteeing the output distribution is exactly $P_{\mathcal{M}}$ regardless of $S$ --- \swift{} is lossless by construction.

\textbf{Online skip-set search.} As with \conflayers{}'s $\tau$, $S$ is searched online, trading a cheaper draft (larger $S$) against a lower acceptance rate,
\begin{equation}
S^\star = \argmin_{S \subseteq \{1,\dots,L\}} \; \mathbb{E}_x\!\left[\frac{\mathrm{WallClockCost}(x; S)}{\mathbb{E}[\text{accepted tokens per round} \mid S]}\right],
\label{eq:swift-search}
\end{equation}
again via Bayesian optimization over the discrete space of skip sets, separated from pure drafting cost in Section~\ref{sec:overhead}.

\section{Experimental Setup}
\label{sec:setup}

\textbf{Models.} We use Qwen2.5-0.5B and Qwen2.5-1.5B \citep{yang2024qwen2} as backbones.

\textbf{Tasks.} GSM8K \citep{cobbe2021gsm8k} (grade-school math word problems, exact-match accuracy) represents multi-step reasoning; CNN/DailyMail \citep{nallapati2016cnndm,see2017cnndm} (news summarization, ROUGE-L, \citealt{lin2004rouge}) represents long-form generation with a different token budget and error profile. We evaluate 100 held-out samples per task, with an identical shuffle/seed recipe across dataset aliases, matching the default evaluation size of the \conflayers{} and \swift{} reference implementations' own scripts. We report per-seed values (Appendix~\ref{sec:appendix-perseed}) rather than a single point estimate, since 100 samples is small enough that seed-to-seed variance should be inspected directly (Section~\ref{sec:discussion}).

\textbf{Baselines.} \emph{Vanilla} is standard autoregressive decoding with no acceleration. \emph{\conflayers} and \emph{\swift} are formalized in Sections~\ref{sec:background-conflayers} and~\ref{sec:background-swift}; \swift{} results use a from-scratch port of the official self-drafting and tree-attention implementation to Qwen2, verified class-by-class against the reference implementation.

\textbf{Rigor protocol.} Every accuracy and cost cell (method $\times$ task $\times$ scale) is run at three seeds (2024, 42, 123), reported as a mean with the observed range. We verified that a CNN/DailyMail generation-length normalization does not itself explain observed score differences between methods.

\subsection{Implementation details}
\label{sec:implementation-details}

\textbf{\conflayers{} search calibration.} We run \conflayers{}'s online threshold search (Equation~\ref{eq:conf-search}) with its reference implementation's default configuration: search interval 30, maximum optimization iterations 100, maximum score threshold 0.95, context window 100 tokens. Thresholds are calibrated online, per query, not per task.

\textbf{\swift{} search calibration.} \swift{}'s online skip-set search (Equation~\ref{eq:swift-search}) uses its reference default: context window 50, optimization interval 1, Bayesian-optimization interval 25, maximum optimization iterations 1000, maximum tolerance iterations 300, maximum score threshold 0.93, with an upper-confidence-bound acquisition function ($\kappa{=}2.5$).

\textbf{Cost measurement protocol.} All latency measurements use batch size 1 on a single GPU, with explicit device synchronization before and after each timed call so reported wall-clock time reflects completed GPU work rather than queued kernels. Reported time includes the full generation call --- sampling, KV-cache updates, any interleaved search --- excluding tokenization/detokenization. We did not run warm-up iterations; the first query of each run carries unsubtracted one-time initialization overhead (Section~\ref{sec:discussion}).

\section{Results}
\label{sec:results}

\subsection{Accuracy}
\label{sec:accuracy}

\begin{figure}[t]
\begin{center}
\includegraphics[width=\linewidth]{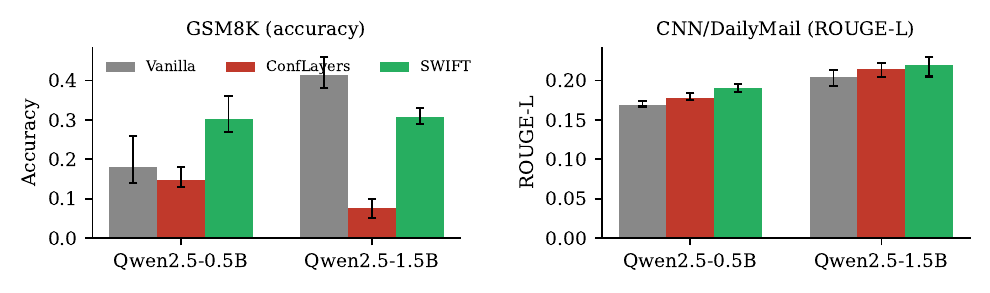}
\end{center}
\caption{Accuracy across the full 2 (task) $\times$ 2 (scale) $\times$ 3 (method) matrix. Error bars show the observed 3-seed range. \conflayers{} is the consistent worst performer in every cell.}
\label{fig:accuracy-matrix}
\end{figure}

Figure~\ref{fig:accuracy-matrix} and Table~\ref{tab:accuracy} summarize the accuracy matrix for vanilla decoding and the two periodic-step methods. The ranking is stable across all three seeds in every cell.

\begin{table}[t]
\caption{Accuracy, mean over 3 seeds (range in brackets). GSM8K is exact-match accuracy; CNN/DailyMail is ROUGE-L. Bold marks the best method per cell.}
\label{tab:accuracy}
\begin{center}
\small
\begin{tabular}{llccc}
\toprule
Task & Scale & Vanilla & \conflayers{} & \swift{} \\
\midrule
\multirow{2}{*}{GSM8K} & 0.5B & 0.180 [0.14--0.26] & 0.147 [0.13--0.18] & \textbf{0.303} [0.27--0.36] \\
 & 1.5B & \textbf{0.413} [0.38--0.46] & 0.077 [0.05--0.10] & 0.307 [0.29--0.33] \\
\midrule
\multirow{2}{*}{CNN/DM} & 0.5B & 0.169 [0.167--0.174] & 0.178 [0.175--0.184] & \textbf{0.190} [0.185--0.196] \\
 & 1.5B & 0.204 [0.193--0.213] & 0.215 [0.204--0.222] & \textbf{0.219} [0.205--0.230] \\
\bottomrule
\end{tabular}
\end{center}
\end{table}

\conflayers{} underperforms both other methods in every condition, with particularly large deficits on GSM8K at 1.5B (0.077 vs. vanilla's 0.413), consistent with prior observations that fixed-representation confidence estimates are unreliable for multi-step reasoning \citep{schuster2022calm}. \swift{} is strongest in three of four cells, losing only to vanilla decoding on GSM8K at 1.5B (0.307 vs. 0.413) --- but still substantially outperforming \conflayers{} there (0.307 vs. 0.077), so the ranking between the two search-based methods is stable across all four cells.

\subsection{Cost}
\label{sec:cost}

\begin{figure}[t]
\begin{center}
\includegraphics[width=\linewidth]{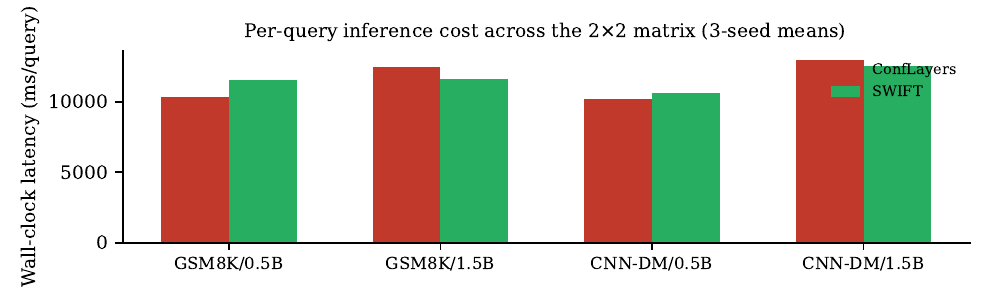}
\end{center}
\caption{Per-query wall-clock latency across the full matrix.}
\label{fig:speed-matrix}
\end{figure}

Table~\ref{tab:cost} and Figure~\ref{fig:speed-matrix} report per-query wall-clock latency for the two periodic-step methods.

\begin{table}[t]
\caption{Per-query wall-clock latency (ms), mean over 3 seeds. Both methods shown decomposed into pure inference and online-search overhead (Section~\ref{sec:overhead}).}
\label{tab:cost}
\begin{center}
\small
\begin{tabular}{llccc}
\toprule
Task/Scale & Method & Pure inference (ms) & Search overhead & Wall-clock total (ms) \\
\midrule
\multirow{2}{*}{GSM8K/0.5B} & \conflayers{} & 9194 & 2.2\% & 10338.5 \\
 & \swift{} & 8771 & 15.1\% & 11564.5 \\
\midrule
\multirow{2}{*}{GSM8K/1.5B} & \conflayers{} & 12366 & 1.3\% & 12485.4 \\
 & \swift{} & 9913 & 12.4\% & 11628.8 \\
\midrule
\multirow{2}{*}{CNN-DM/0.5B} & \conflayers{} & 10889 & 1.2\% & 10231.4 \\
 & \swift{} & 8610 & 8.7\% & 10598.2 \\
\midrule
\multirow{2}{*}{CNN-DM/1.5B} & \conflayers{} & 12555 & 1.2\% & 12973.4 \\
 & \swift{} & 9880 & 18.6\% & 12532.5 \\
\bottomrule
\end{tabular}
\end{center}
\end{table}

\subsubsection{Search-overhead decomposition}
\label{sec:overhead}

\begin{figure}[t]
\begin{center}
\includegraphics[width=\linewidth]{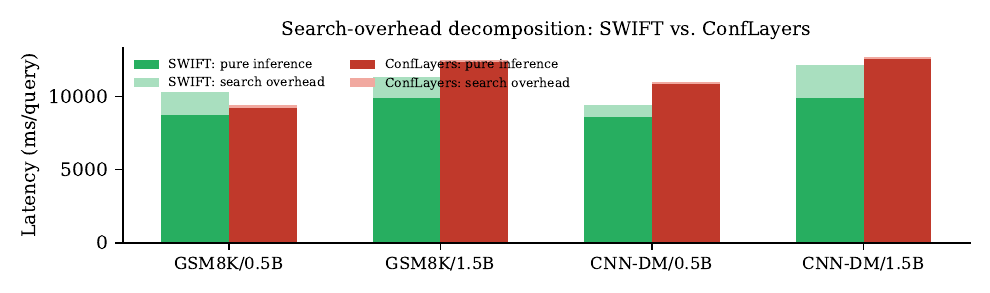}
\end{center}
\caption{Search-overhead decomposition. Once online-search cost is separated from pure model-forward cost, \swift{} is the faster inference engine in every cell, despite appearing slower on unsplit wall-clock time in three of four cells.}
\label{fig:search-overhead}
\end{figure}

Naive wall-clock comparison ranks \conflayers{} as faster than \swift{} in three of four cells (Table~\ref{tab:cost}, ``wall-clock total''). This is misleading. We instrumented both methods to separately log pure model-forward-pass time and the wall-clock time consumed by each method's online search --- Bayesian-optimization threshold search for \conflayers{}, and the analogous search over which layers to self-draft-skip for \swift{}. Figure~\ref{fig:search-overhead} shows the result: once decomposed, \swift{}'s pure inference cost is \emph{faster} than \conflayers{}'s in all four cells (5--21\%), reversing the naive ranking in three of them. \conflayers{}'s search overhead is small and stable (1.2--2.2\% of total cost, near-zero variance across seeds); \swift{}'s is larger (8.7--18.6\% mean) and more variable (up to 28.7\% on individual GSM8K/0.5B seeds).

Every prior cost comparison involving \swift{} in this line of work understated its true inference speed by folding search cost into the reported number. This decomposition assumes many queries are served within one running session: both methods' search state persists across queries in our harness rather than resetting per query, so search overhead is front-loaded onto early queries and shrinks as the search converges. We recommend future efficiency comparisons report this decomposition explicitly, alongside a total-cost-for-$N$-queries view when $N$ is known, rather than a single averaged wall-clock figure.

\section{Supplemental Analysis: Trained Routing at Coarser Granularity}
\label{sec:supplemental}

This section examines \layerroute{} and \layerdrop{}, two trained-routing methods operating at a coarser decision granularity than \conflayers{} and \swift{} (Section~\ref{sec:taxonomy}), not as direct competitors to the main comparison but under a measurement protocol verified to exercise genuine per-input gating, a genuine full-model baseline, and genuine inference-time compute skipping (Section~\ref{sec:supplemental-layerroute}).

\subsection{Method formalizations}
\label{sec:supplemental-formalizations}

\textbf{\layerroute{} \citep{sikdar2026layerroute}.} \layerroute{} freezes a pretrained backbone $\theta$ and attaches, at every transformer layer $l \in \{1, \dots, L\}$, a low-rank adapter $\Delta W_l$ \citep{hu2021lora} with a scalar gate head. For hidden state $h_{l-1}$ entering layer $l$,
\begin{equation}
f_l(h_{l-1}) = h_{l-1} + \mathrm{Attn}_l(h_{l-1}) + \mathrm{FFN}_{\Delta W_l}(h_{l-1}),
\label{eq:layer}
\end{equation}
and the gate head produces a scalar logit converted to a hard binary decision
\begin{equation}
z_l = \mathbb{1}\left[\sigma\!\left(g_l(h_{l-1})\right) > 0.5\right], \qquad \sigma(x) = \frac{1}{1+e^{-x}},
\label{eq:gate}
\end{equation}
with layer output $h_l = z_l \, f_l(h_{l-1}) + (1 - z_l)\, h_{l-1}$, gradients propagated via straight-through estimation \citep{bengio2013estimating}. The gate head pools over the \emph{current full sequence's} hidden states, making this a per-sequence decision (Table~\ref{tab:taxonomy}).\footnote{\layerroute{}'s original release reported a skip-rate differential between tool-call and planning inputs; this did not reproduce under multi-seed testing and has since been withdrawn by its authors. The measurement protocol and results in this section were obtained under the corrected training recipe throughout and do not depend on that claim.} All parameters --- router and LoRA --- are trained jointly against a single gate-regularized language-modeling objective,
\begin{equation}
\mathcal{L} = \mathcal{L}_{\mathrm{LM}} + \lambda \cdot \frac{1}{L}\sum_{i} \sigma(s_i),
\label{eq:gatereg}
\end{equation}
with $\lambda = 1.0$ discouraging collapse to an all-open equilibrium; no auxiliary classification signal (e.g. tool-call/planning provenance) is used. Schematic in Figure~\ref{fig:arch-layerroute}.

\textbf{\layerdrop{} \citep{fan2020layerdrop}.} We adapt \layerdrop{} to the same LoRA budget as \layerroute{} for a matched training-cost comparison. During training, each layer is stochastically dropped with probability $p$ (uniform across layers), training the backbone and LoRA adapters to be robust to any layer's removal. At inference, a \emph{fixed} set of layers, chosen once from training-time drop statistics, is pruned identically for every input --- no router, no per-input adaptivity, the coarsest regime in Table~\ref{tab:taxonomy}. Schematic in Figure~\ref{fig:arch-layerdrop}.

\subsection{Measurement protocol}
\label{sec:supplemental-layerroute}

Because \layerroute{}'s architecture blends a gated and an ungated output rather than branching between them, measuring it correctly requires care on three points, each verified independently.

\textbf{Genuine gating.} Wall-clock cost is measured through \layerroute{}'s gated forward pass directly, token by token, verified to produce output that genuinely differs from an ungated pass rather than being byte-identical to it.

\textbf{Genuine full-model baseline.} The baseline forces every layer open via direct bias manipulation on the gate head, verified on a controlled synthetic test to change the layer-execution count to the full layer count.

\textbf{Genuine inference-time compute skipping.} Training computes every layer's full transformation unconditionally (a necessary property of the straight-through estimator, whose backward pass needs the layer's real forward value even for a gate evaluating to zero); inference adds a conditional skip where no such constraint applies (\texttt{if gate==0 and not self.training: continue}), verified via layer-call counting to produce fewer real invocations than the total layer count, with output identical to the blended-but-discarded version for closed gates.

For long-form generation, we verified the generation loop's stopping condition is decoupled from prompt-length truncation, since CNN/DailyMail articles can approach the truncation limit; the post-hoc token-count distribution matches the expected range for both methods, with no queries terminating after a single token.

\layerdrop{}'s fixed pattern requires only the baseline and skip verifications, having no per-input gate to exercise. \conflayers{} and \swift{}'s results (Sections~\ref{sec:results}--\ref{sec:overhead}) use independent evaluation code, verified separately under the same standard.

\subsection{Accuracy and cost}
\label{sec:supplemental-results}

\begin{figure}[t]
\begin{center}
\includegraphics[width=\linewidth]{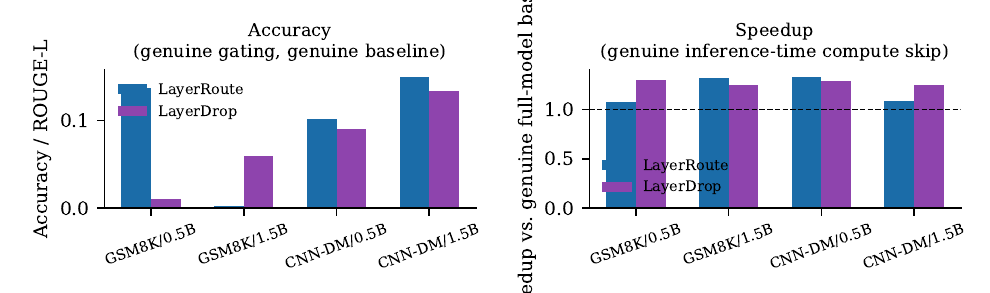}
\end{center}
\caption{Supplemental results. Left: accuracy under genuine gating and a genuine full-model baseline. Right: speedup versus that same genuine baseline, including genuine inference-time compute skipping (Section~\ref{sec:supplemental-layerroute}).}
\label{fig:supplemental-matrix}
\end{figure}

Table~\ref{tab:supplemental} and Figure~\ref{fig:supplemental-matrix} report accuracy and speedup, 3 seeds each, using a genuine full-model baseline throughout.

\begin{table}[t]
\caption{Supplemental analysis: accuracy and speedup, mean over 3 seeds (range in brackets), \layerroute{} and \layerdrop{} versus a genuine full-model baseline. GSM8K is exact-match; CNN/DailyMail is ROUGE-L. Speedup $>1$ indicates the routed/pruned pass is faster than the full-model baseline.}
\label{tab:supplemental}
\begin{center}
\small
\begin{tabular}{llcccc}
\toprule
Task & Scale & \layerroute{} acc. & \layerroute{} speedup & \layerdrop{} acc. & \layerdrop{} speedup \\
\midrule
\multirow{2}{*}{GSM8K} & 0.5B & 0.137 [0.11--0.16] & 1.08$\times$ & 0.010 [0.00--0.02] & 1.30$\times$ \\
 & 1.5B & 0.003 [0.00--0.01] & 1.32$\times$ & 0.060 [0.05--0.07] & 1.25$\times$ \\
\midrule
\multirow{2}{*}{CNN/DM} & 0.5B & 0.102 [0.099--0.107] & 1.33$\times$ & 0.091 [0.085--0.097] & 1.29$\times$ \\
 & 1.5B & 0.150 [0.139--0.165] & 1.09$\times$ & 0.134 [0.131--0.136] & 1.25$\times$ \\
\bottomrule
\end{tabular}
\end{center}
\end{table}

Two findings stand out. First, \layerroute{}'s accuracy on GSM8K at 1.5B is essentially zero (0.003 mean, 1 of 300 samples correct) --- a near-total collapse. Routed generations at this scale are coherent, grammatically well-formed text, not degenerate repetition, but consistently fail to reach a correct answer, which we interpret as a genuine capability gap under real per-sequence gating. Second, both methods show modest but real speedups (1.08--1.33$\times$) over a genuine full-model baseline, smaller than layer-count-implied theoretical savings would suggest; we attribute this to \layerroute{}'s router computation (pooling and a linear projection, evaluated at every layer regardless of that layer's decision) contributing fixed per-layer overhead. \layerdrop{}, having no such per-query overhead, shows a somewhat larger, more consistent speedup (1.25--1.30$\times$) despite lower average accuracy.

These results should not be read as ``trained routing underperforms search-based routing'' in general: \layerroute{} and \layerdrop{} operate at a different decision granularity (Section~\ref{sec:taxonomy}), were trained under a fixed, unremarkable LoRA budget not tuned per condition, and represent two specific instances of a much larger design space.

\section{Discussion and Limitations}
\label{sec:discussion}

\textbf{Main finding.} Among periodic-step, search-based methods, \swift{} is stronger on both accuracy and (once search overhead is decomposed) speed; \conflayers{} is dominated on both axes throughout, stable across all seeds in every cell.

\textbf{On the supplemental analysis.} \layerroute{} and \layerdrop{} occupy a coarser, less input-adaptive regime than \conflayers{}/\swift{} (Section~\ref{sec:taxonomy}); the honest conclusion is that under a fixed, unremarkable LoRA budget, and measured under the verified protocol of Section~\ref{sec:supplemental-layerroute}, neither is competitive on accuracy here, while both show modest, genuine speedups once compute skipping is confirmed to actually occur.

\textbf{Scope of the audit.} We evaluate two model scales and two tasks on instruction-tuned backbones only; findings may not hold for base models. We follow the reference implementations' default evaluation size (100 samples) without a power analysis for confidence-interval width. Given $n{=}3$ seeds, we report per-seed values directly (Appendix~\ref{sec:appendix-perseed}) rather than a higher-powered parametric test, and we do not run warm-up iterations before timing. Memory footprint, integration complexity, distribution-shift robustness, and combinability with techniques such as quantization are also relevant to deployment but outside this audit's scope.

\subsection*{AI Use Statement}
Generative AI tools assisted with literature survey (verified against original sources) and evaluation scaffolding, including the search-overhead decomposition harness and plotting utilities (tested against known reference outputs). All experimental design, method choices, and interpretation of results are the author's own, who takes responsibility for this work's final content.

\subsection*{Reproducibility Statement}
Figures in Sections~\ref{sec:results}--\ref{sec:supplemental} are means over three fixed seeds (2024, 42, 123) with per-seed values in Appendix~\ref{sec:appendix-perseed}; full protocol in Sections~\ref{sec:setup}--\ref{sec:implementation-details} and~\ref{sec:supplemental-layerroute}. The \swift{} port was verified class-by-class against its upstream implementation. We intend to release our harness, the Qwen2 \swift{} port, trained checkpoints, and per-query result files with this paper.

\bibliography{refs}
\bibliographystyle{iclr2027_conference}

\clearpage
\appendix
\section{Architecture Diagrams}
\label{app:diagrams}

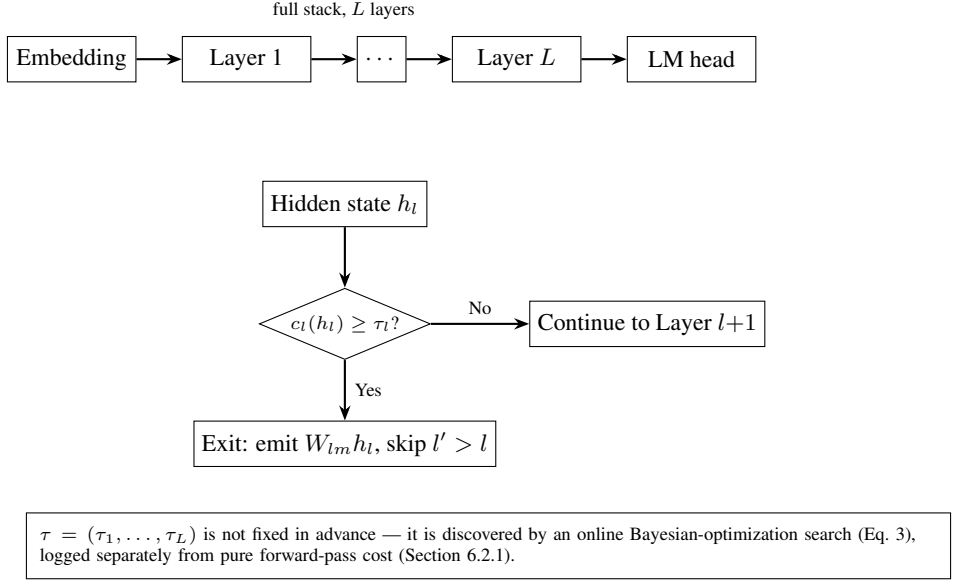
\begin{figure}[H]
\begin{center}
\begin{tikzpicture}[node distance=6mm and 6mm]
\node[box] (emb) {Embedding};
\node[box, right=of emb] (l1) {Layer 1};
\node[box, right=of l1, minimum width=0.5cm] (dots) {$\cdots$};
\node[box, right=of dots] (lL) {Layer $L$};
\node[box, right=of lL] (lmh) {LM head};
\draw[arr] (emb) -- (l1);
\draw[arr] (l1) -- (dots);
\draw[arr] (dots) -- (lL);
\draw[arr] (lL) -- (lmh);
\node[font=\scriptsize, above=1mm of emb, xshift=3.6cm] {full stack, $L$ layers};

\node[box, below=13mm of l1, xshift=1.3cm] (hl) {Hidden state $h_l$};
\node[decision, below=8mm of hl] (dec) {$c_l(h_l) \ge \tau_l$?};
\node[box, right=13mm of dec] (cont) {Continue to Layer $l{+}1$};
\node[box, below=8mm of dec] (exit) {Exit: emit $W_{lm}h_l$, skip $l' > l$};
\draw[arr] (hl) -- (dec);
\draw[arr] (dec) -- node[above,font=\scriptsize]{No} (cont);
\draw[arr] (dec) -- node[right,font=\scriptsize]{Yes} (exit);
\node[note, below=6mm of exit, xshift=1.9cm] {$\tau = (\tau_1,\dots,\tau_L)$ is not fixed in advance --- it is discovered by an online Bayesian-optimization search (Eq.~3), logged separately from pure forward-pass cost (Section~\ref{sec:overhead}).};
\end{tikzpicture}
\end{center}
\caption{\conflayers{}: top, the full layer stack; bottom, detail of the per-layer confidence check. A per-layer confidence estimate $c_l$ is thresholded against $\tau_l$, discovered online via Bayesian optimization; pure inference and search overhead are logged and reported separately (Section~\ref{sec:overhead}).}
\label{fig:arch-conflayers}
\end{figure}

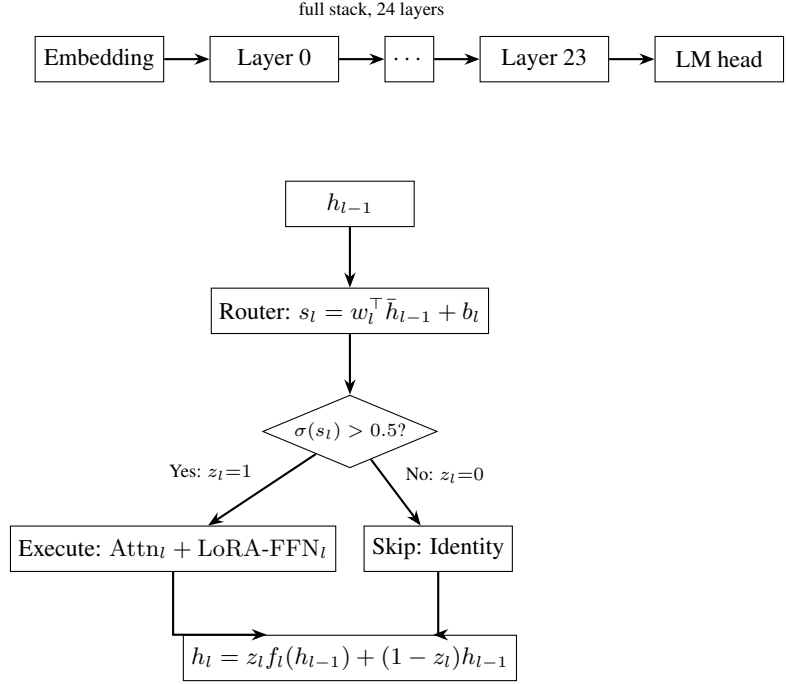
\begin{figure}[H]
\begin{center}
\begin{tikzpicture}[node distance=6mm and 6mm]
\node[box] (emb) {Embedding};
\node[box, right=of emb] (l0) {Layer 0};
\node[box, right=of l0, minimum width=0.5cm] (dots) {$\cdots$};
\node[box, right=of dots] (l23) {Layer 23};
\node[box, right=of l23] (lmh) {LM head};
\draw[arr] (emb) -- (l0);
\draw[arr] (l0) -- (dots);
\draw[arr] (dots) -- (l23);
\draw[arr] (l23) -- (lmh);
\node[font=\scriptsize, above=1mm of emb, xshift=3.6cm] {full stack, 24 layers};

\node[box, below=13mm of l0, xshift=1cm] (hl) {$h_{l-1}$};
\node[box, below=8mm of hl] (router) {Router: $s_l = w_l^\top \bar h_{l-1} + b_l$};
\node[decision, below=8mm of router] (dec) {$\sigma(s_l) > 0.5$?};
\node[box, below left=10mm and -4mm of dec] (exec) {Execute: $\mathrm{Attn}_l + \mathrm{LoRA\text{-}FFN}_l$};
\node[box, below right=10mm and -4mm of dec] (skip) {Skip: Identity};
\node[box, below=22mm of dec] (merge) {$h_l = z_l f_l(h_{l-1}) + (1-z_l) h_{l-1}$};

\draw[arr] (hl) -- (router);
\draw[arr] (router) -- (dec);
\draw[arr] (dec) -- node[above left,font=\scriptsize]{Yes: $z_l{=}1$} (exec);
\draw[arr] (dec) -- node[above right,font=\scriptsize]{No: $z_l{=}0$} (skip);
\draw[arr] (exec.south) |- ($(merge.north)+(-1.1cm,0)$);
\draw[arr] (skip.south) |- ($(merge.north)+(1.1cm,0)$);
\end{tikzpicture}
\end{center}
\caption{\layerroute{}: top, the full 24-layer stack; bottom, detail of a single gated layer $l$. The router computes a scalar score from the mean-pooled input hidden state $\bar h_{l-1}$, thresholds it into a hard gate $z_l$ via a straight-through estimator, and the layer output is either the LoRA-adapted block's output ($z_l{=}1$) or an unchanged residual pass-through ($z_l{=}0$), matching Equations~\ref{eq:layer}--\ref{eq:gate}. Discussed in the supplemental analysis, Section~\ref{sec:supplemental}.}
\label{fig:arch-layerroute}
\end{figure}

Figures~\ref{fig:arch-conflayers} and~\ref{fig:arch-swift} give schematic diagrams for the two main-comparison methods; Figures~\ref{fig:arch-layerroute} and~\ref{fig:arch-layerdrop} diagram \layerroute{} and \layerdrop{}, discussed in the supplemental analysis (Section~\ref{sec:supplemental}).

\clearpage

\begin{figure}[H]
\begin{center}
\begin{tikzpicture}[node distance=4mm and 16mm]
\node[font=\scriptsize\bfseries] (hdrL) {Self-draft $M_S$};
\node[box, below=6mm of hdrL] (dl1) {Layer 1: used};
\node[box, below=of dl1] (dl2) {Layer 2: used};
\node[dropbox, below=of dl2] (dl3) {Layer 3: skipped};
\node[dropbox, below=of dl3] (dl4) {Layer 4: skipped};
\node[box, below=of dl4] (dl5) {Layer 5: used};
\draw[arr] (dl1)--(dl2); \draw[arr] (dl2)--(dl3); \draw[arr] (dl3)--(dl4); \draw[arr] (dl4)--(dl5);

\node[font=\scriptsize\bfseries, right=16mm of hdrL] (hdrR) {Verify $M$ (full backbone)};
\node[box, below=6mm of hdrR] (vl1) {Layer 1: verify};
\node[box, below=of vl1] (vl2) {Layer 2: verify};
\node[box, below=of vl2] (vl3) {Layer 3: verify};
\node[box, below=of vl3] (vl4) {Layer 4: verify};
\node[box, below=of vl4] (vl5) {Layer 5: verify};
\draw[arr] (vl1)--(vl2); \draw[arr] (vl2)--(vl3); \draw[arr] (vl3)--(vl4); \draw[arr] (vl4)--(vl5);

\node[box, below=10mm of dl5, xshift=8mm, minimum width=6.5cm] (accept) {Accept longest verified prefix (rejection sampling, Eq.~5)};
\draw[arr] (dl5.south) |- ($(accept.north)+(-1.4cm,0)$);
\draw[arr] (vl5.south) |- ($(accept.north)+(1.4cm,0)$);
\node[note, below=6mm of accept, xshift=0cm] {$S$, the set of skipped layers, is searched online (Eq.~6) to trade a cheaper draft against a lower acceptance rate --- logged separately from pure drafting cost (Section~\ref{sec:overhead}).};
\end{tikzpicture}
\end{center}
\caption{\swift{}: the target model self-drafts by skipping a searched subset $S$ of its own layers, then the full model verifies all draft positions in a single batched pass; the longest accepted prefix is kept and the rest resampled.}
\label{fig:arch-swift}
\end{figure}

\clearpage

\clearpage

\begin{figure}[H]
\begin{center}
\begin{tikzpicture}[node distance=6mm and 6mm]
\node[box] (emb) {Embedding};
\node[box, right=of emb] (l1) {Layer 1: kept};
\node[dropbox, right=of l1] (l2) {Layer 2: dropped};
\node[box, right=of l2] (l3) {Layer 3: kept};
\node[box, right=of l3, minimum width=0.5cm] (dots) {$\cdots$};
\node[dropbox, right=of dots] (lL) {Layer $L{-}1$: dropped};
\node[box, right=of lL] (lmh) {LM head};
\draw[arr] (emb)--(l1); \draw[arr] (l1)--(l2); \draw[arr] (l2)--(l3);
\draw[arr] (l3)--(dots); \draw[arr] (dots)--(lL); \draw[arr] (lL)--(lmh);

\node[note, below=10mm of l2, xshift=1.6cm, text width=0.8\linewidth] (n1) {\textbf{Fixed, input-independent pattern.} Which layers are kept versus dropped is chosen once from training-time drop statistics (each layer stochastically dropped with probability $p$ during training) and then applied \emph{identically to every input} at inference --- no router, no per-input adaptivity, the coarsest regime in Table~\ref{tab:taxonomy}.};
\end{tikzpicture}
\end{center}
\caption{\layerdrop{}: layers are stochastically dropped during training to make the backbone and LoRA adapters robust to any layer's removal; at inference, one fixed set of layers, chosen once from training statistics, is pruned identically for every input --- no router, no per-input adaptivity. Discussed in the supplemental analysis, Section~\ref{sec:supplemental}.}
\label{fig:arch-layerdrop}
\end{figure}

\clearpage

\section{Additional Per-Seed Results}
\label{sec:appendix-perseed}
\label{app:perseed}

\subsection{Main comparison: ConfLayers and SWIFT}

Table~\ref{tab:accuracy} in the main text reports each accuracy cell as a mean over three seeds with the observed range. Tables~\ref{tab:perseed-gsm8k} and~\ref{tab:perseed-cnndm} give every individual per-seed value.

\begin{table}[h]
\caption{GSM8K exact-match accuracy at each of the three evaluation seeds, main comparison.}
\label{tab:perseed-gsm8k}
\begin{center}
\small
\begin{tabular}{llccc}
\toprule
Scale & Seed & Vanilla & \conflayers{} & \swift{} \\
\midrule
\multirow{3}{*}{0.5B} & 2024 & 0.26 & 0.13 & 0.36 \\
 & 42 & 0.14 & 0.18 & 0.28 \\
 & 123 & 0.14 & 0.13 & 0.27 \\
\midrule
\multirow{3}{*}{1.5B} & 2024 & 0.40 & 0.05 & 0.29 \\
 & 42 & 0.46 & 0.08 & 0.33 \\
 & 123 & 0.38 & 0.10 & 0.30 \\
\bottomrule
\end{tabular}
\end{center}
\end{table}

\begin{table}[h]
\caption{CNN/DailyMail ROUGE-L at each of the three evaluation seeds, main comparison.}
\label{tab:perseed-cnndm}
\begin{center}
\small
\begin{tabular}{llccc}
\toprule
Scale & Seed & Vanilla & \conflayers{} & \swift{} \\
\midrule
\multirow{3}{*}{0.5B} & 2024 & 0.167 & 0.175 & 0.190 \\
 & 42 & 0.174 & 0.184 & 0.196 \\
 & 123 & 0.167 & 0.178 & 0.185 \\
\midrule
\multirow{3}{*}{1.5B} & 2024 & 0.193 & 0.204 & 0.205 \\
 & 42 & 0.206 & 0.218 & 0.223 \\
 & 123 & 0.213 & 0.222 & 0.230 \\
\bottomrule
\end{tabular}
\end{center}
\end{table}

\subsection{Supplemental analysis: LayerRoute and LayerDrop}

Tables~\ref{tab:perseed-supp-gsm8k} and~\ref{tab:perseed-supp-cnndm} give per-seed values underlying Table~\ref{tab:supplemental}, measured under the protocol described in Section~\ref{sec:supplemental-layerroute}.

\begin{table}[h]
\caption{GSM8K exact-match accuracy at each of the three evaluation seeds, supplemental analysis.}
\label{tab:perseed-supp-gsm8k}
\begin{center}
\small
\begin{tabular}{llcc}
\toprule
Scale & Seed & \layerroute{} & \layerdrop{} \\
\midrule
\multirow{3}{*}{0.5B} & 2024 & 0.14 & 0.00 \\
 & 42 & 0.16 & 0.02 \\
 & 123 & 0.11 & 0.01 \\
\midrule
\multirow{3}{*}{1.5B} & 2024 & 0.00 & 0.07 \\
 & 42 & 0.01 & 0.05 \\
 & 123 & 0.00 & 0.06 \\
\bottomrule
\end{tabular}
\end{center}
\end{table}

\begin{table}[h]
\caption{CNN/DailyMail ROUGE-L at each of the three evaluation seeds, supplemental analysis.}
\label{tab:perseed-supp-cnndm}
\begin{center}
\small
\begin{tabular}{llcc}
\toprule
Scale & Seed & \layerroute{} & \layerdrop{} \\
\midrule
\multirow{3}{*}{0.5B} & 2024 & 0.100 & 0.085 \\
 & 42 & 0.099 & 0.092 \\
 & 123 & 0.107 & 0.097 \\
\midrule
\multirow{3}{*}{1.5B} & 2024 & 0.146 & 0.136 \\
 & 42 & 0.139 & 0.131 \\
 & 123 & 0.165 & 0.134 \\
\bottomrule
\end{tabular}
\end{center}
\end{table}

\end{document}